\documentclass[a4paper]{article}

\usepackage{INTERSPEECH2018}
\usepackage{CJKutf8}
\usepackage[]{algorithmic}
\title{Study of Semi-supervised Approaches to Improving English-Mandarin Code-Switching Speech Recognition}
\name{Pengcheng Guo$^{1,2}$, Haihua Xu$^2$, Lei Xie\thanks{$*$ Lei Xie is the corresponding author.}$^{1,*}$, Eng Siong Chng$^{2,3}$}
\address{
  $^1$ School of Computer Science, Northwestern Polytechnical University, Xi'an, China \\
  $^2$ Temasek Laboratories, Nanyang Technological University, Singapore \\
  $^3$ School of Computer Science and Engineering, Nanyang Technological University, Singapore
  }
 \email{guopengcheng1220@gmail.com, lxie@nwpu.edu.cn, \{haihuaxu,aseschng\}@ntu.edu.sg}

\begin{document}
\maketitle
\begin{abstract}

In this paper, we present our overall efforts to improve the performance of a code-switching speech recognition system using semi-supervised training methods from lexicon learning to acoustic modeling, on the South East Asian Mandarin-English (SEAME) data.
We first investigate semi-supervised lexicon learning approach
to adapt the canonical lexicon, which is meant to alleviate the heavily accented pronunciation issue within the code-switching conversation of the local area. 
As a result, the learned lexicon yields improved performance.
 Furthermore, we attempt to use semi-supervised training to deal with those transcriptions that are highly mismatched between human transcribers and ASR system.
Specifically, we conduct semi-supervised training assuming those poorly transcribed data as unsupervised data. We found the semi-supervised acoustic modeling can lead to improved results.
Finally, to make up for the limitation of the conventional n-gram language models due to 
data sparsity issue, we perform lattice rescoring using neural network language models, and significant WER reduction is obtained.
\end{abstract}
\noindent\textbf{Index Terms}: speech recognition, code-switching, lexicon learning, semi-supervised training, lattice rescoring

\section{Introduction} \label{sec:intro}
In recent years, code-switching speech recognition research has been drawn increasing attention in speech recognition community. This is because most existing state-of-the-art speech recognition engines are only capable of understanding a specific monolingual language, say English or Mandarin. When people converse with mixed language simultaneously, for instance, \begin{CJK*}{UTF8}{gbsn}``请问{ }到{ }changi airport{ }怎么{ }走?''\end{CJK*}, monolingual speech recognition engines failed for the code-switching part. This is common in South East Asia area, where people can usually speak several languages at the same time and code-switching conversation is very common \cite{auer2013code}.

Compared with the building of a monolingual speech recognition system, it is much more challenging to build a code-switching speech recognition system, since one is often encountered with data sparsity problem for either individual languages. 

First, to address data sparsity issue in acoustic modeling, cross-lingual phone merging \cite{mak1996phone,li2011asymmetric,lin2009study} even global phone set usage \cite{Association1999Handbook} was widely studied in the era of using GMM-HMM acoustic modeling method. This indeed enforced cross-lingual data sharing and alleviated the intensity of individual language data sparsity issue to some degree, but it also brought about cross-lingual confusion problem. That is, inter-language substitution issue arises due to the discriminative capability of the merged phone being undermined. The underlying reason is that there are really some phones that are
acoustically similar but their acoustic contexts are significantly different
in each individual languages. 
However, as the advent of deep neural network (DNN) acoustic modeling techniques, the necessity of cross-lingual phone merging or global phone usage is decreased thanks to the topology of DNN architecture. Since the lower hidden layers are inherently shared among different languages and can be jointly learned by each individual languages \cite{huang2013cross,yilmaz2016investigating}.

Secondly, it is also challenging to robustly estimate language models for code-switching speech recognition. Take the conventional bigram language models as an example.  It is easy to access sufficient counts for the bigrams within each monolingual languages; however it is hard to get those cross-lingual bigram counts, such as \begin{CJK*}{UTF8}{gbsn} ``到{ }changi'' or ``airport 怎么'' \end{CJK*} etc. as is seen from the above exemplar utterance. 
On the one hand, one can employ machine translation method to generate a lot of artificial code-switching text utterances \cite{vu2012first}. However limited performance improvement was obtained due to a lot of noisy cross-lingual n-gram counts being introduced.
On the other hand, one can use neural networks based continuous language modeling method, which learns n-gram probability in continuous space and is more powerful to generalize \cite{mikolov2010recurrent,adel2013recurrent,adel2013combination}.

In this paper, we put efforts into improving code-switching speech recognition with semi-supervised training approaches, assuming the imperfection of the original lexicon and human acoustic transcriptions. 

For the lexicon problem, since there is no off-the-shelf South East Asian English lexicon available, we use the CMU open source lexicon to start with. Obviously, there is pronunciation mismatch problem, and  out-of-vocabulary (OOV) problem as well. We first collect alternative pronunciations from original lexicon, G2P and phonetic decoding. Then we adopt a data likelihood based criterion to select and prune all pronunciation candidates.


Once the lexicon is learned, we fix it and analyze the transcription quality. Word Matched Error Rate (WMER) is computed to score the human transcribed transcriptions and the ASR decode hypotheses as in \cite{bell2015mgb}. We found the WMER is around $10\%$ on the SEAME training data, and there is about $7$ hours of $>\!\!30\%$ WMER data.
To deal with these data, we propose to use Lattice-free Maximum Mutual Information (LF-MMI) based semi-supervised training method. We found with semi-supervised training, the poorly transcribed data can also improve the system performance.

This paper is organized as follows: Section \ref{sec:data-desp} describes the distribution of the overall SEAME data. Section \ref{sec:exp-setup} reports the experimental setup. Section \ref{sec:lex} proposes our approach to semi-supervised lexicon learning. Section \ref{sec:semi-am} is for semi-supervised acoustic modeling. Section \ref{sec:lat-rescore} presents neural network language models based lattice rescoring and we conclude in Section \ref{sec:con}.

\section{Data} \label{sec:data-desp}

\subsection{Overall description}

\begin{figure}[htb]
  \centering
  \includegraphics[width=\linewidth]{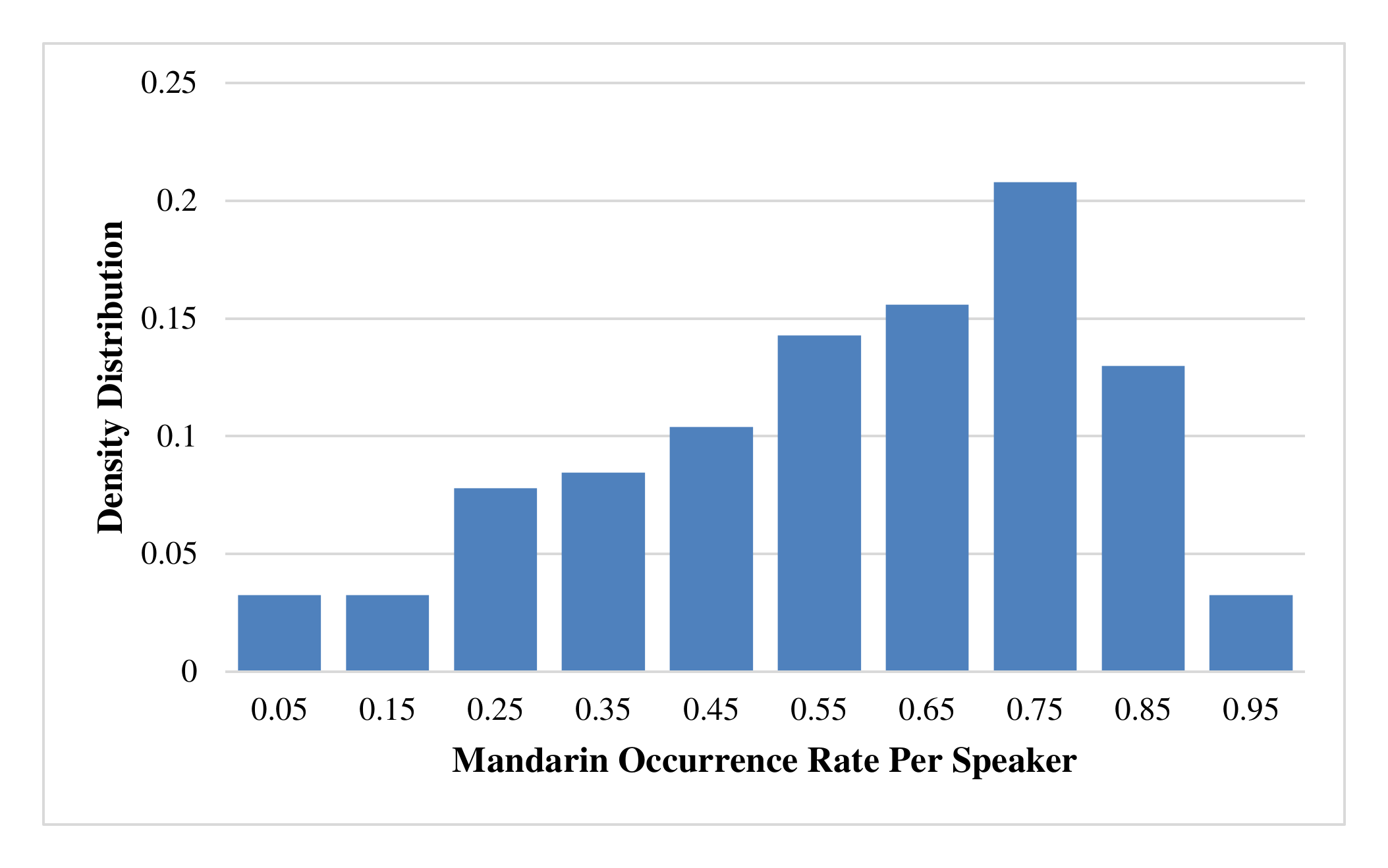}
  \caption{Distribution of speakers versus Mandarin occurrence rates per speaker for the overall SEAME data. Each bar's statistics come from the $\pm 0.05$ range of the corresponding data on the horizontal axis.}
  \label{fig:data}
\end{figure}

The SEAME corpus is a microphone based spontaneous conversational bilingual speech corpus, of which most of the utterances contain both English and Mandarin uttered by 154 oversea Chinese in Malaysia and Singapore areas \cite{vu2012first,lyu2010analysis}. Figure \ref{fig:data} describes the distribution of speakers with regard to Mandarin occurrence rate per speaker. From Figure \ref{fig:data}, we can see the SEAME corpus is generally biased with Mandarin. 

Specifically, about 90\% speakers speak code-switching utterances with Mandarin rate ranged from 10\% to 90\%, while about 2\% speakers have utterances with Mandarin rates no more than 10\%, and 5\% speakers have utterances over 90\% being Mandarin. Besides, we find Singaporean speakers  normally have more English words in their utterances, while Malaysian speakers are more likely to converse with utterances dominated by Mandarin. 

\subsection{Test data definition}
To evaluate the proposed methods, we define two evaluation data sets and each are randomly selected from
10 gender balanced speakers. However they are defined differently, and one is dominated by Mandarin, named as $eval_{man}$, and the other is dominated by English, as $eval_{sge}$. The detailed
statistics are revealed in Table \ref{tab:data-stat}. 
We think these ``biased'' data sets would give more clues to show the effectiveness of each proposed methods on each individual languages.

\begin{table}[htb]
  \caption{ The statistics of two evaluation data sets, one is dominated by Mandarin, named $eval_{man}$, another is dominated by English, named $eval_{sge}.$}
  \label{tab:data-stat}
  \centering
  \begin{tabular}{cccc}
    \toprule
     & Train& $eval_{man}$ & $eval_{sge}$ \\
    \midrule
    Speakers  & $134$   & $10$  & $10$ \\
    Duration (hrs)  & $101.1$  & $7.5$  & $4$   \\
    Mandarin (\%)  & $0.59$   & $0.69$      & $0.29$ \\
    \bottomrule
  \end{tabular}
\end{table}

\section{Experimental setup} \label{sec:exp-setup}

\subsection{Acoustic modeling}
We use Time-delay Neural Networks (TDNNs) \cite{waibel1989modular,peddinti2015time} trained with  lattice-free MMI (LF-MMI) \cite{povey2012generating,povey2016purely} to build acoustic models in all experiments. The front-ends are made up of 40-dimensional MFCC plus 100-dimensional i-vectors \cite{saon2013speaker}. They are LDA transformed over $\pm 2$ concatenated MFCC features plus i-vectors before fed to the TDNN. For  training, the supervisions obtained from the senones of the prior trained GMM-HMM acoustic models.

\subsection{Lexicon and language modeling}
The original lexicon is composed of a CMU English lexicon \cite{cmudict} and a Mandarin lexicon \cite{hsiao2008cmu}. Since we do not merge cross-language phonemes, we separate the two phoneme sets by adding a suffix string of language identity for each phoneme to differentiate in our lexicon. Overall, we have 252 phonemes, in which there are 213 Mandarin and 39 English phonemes respectively. For language modeling, only the transcriptions of the training part of the SEAME data are employed.   

\subsection{Baseline results}
We report the baseline results on the two evaluation sets as defined in Table \ref{tab:data-stat}. To prepare the data, we normalize the overall 
data as follows:
 1) remove those training data that contains $\langle unk \rangle$ words; 2) manually correct about 400 obvious word typos from the human transcriptions. We then train the TDNN acoustic models using the LF-MMI frame sub-sampling method, achieving the WERs of 23.39\% on $eval_{man}$ and 33.04\% on $eval_{sge}$ respectively.
All experiments are based on the Kaldi toolkit \cite{povey2011kaldi}. 
 

\section{Semi-supervised lexicon learning} \label{sec:lex}
\subsection{Motivation} \label{sub:lex-motivation}
As mentioned, our available lexicon is composed of the standard American English and Chinese lexicons, which are mismatched with the SEAME acoustic data, particularly for the English pronunciation part. The mismatched problems are due to two reasons. One is that the large amount of local spoken words leads to a high OOV rate on the SEAME data, about $1.10\%$.
Another reason is that the English pronunciation of the SEAME data is different from the American English, for instance, word ``three'' is normally pronounced as ``tree'' in the SEAME data. 
In this section, we propose to use the supervised G2P training and the unsupervised phonetic decoding to address the mismatched problems, which named semi-supervised lexicon learning.

\subsection{Method} \label{sub:lex-method}
We first address the OOV problem using the G2P method \cite{bisani2008joint}. As a result, we obtain a OOV-free lexicon though pronunciation mismatched problem persists. To solve the pronunciation mismatched problem, we employ a phonetic level decoding process on the training data as inspired from \cite{laurent2010acoustics}. We try to learn adapted word pronunciations using the time boundary of the corresponding word and decoded phone sequence. As this unsupervised procedure would introduce noisy pronunciations for some words, pronunciation probability estimation and pruning is crucial. We adopt the estimation and pruning approaches proposed in \cite{zhang2017acoustic}, and the whole procedure of the method is depicted in Figure \ref{fig:lex-learn}. 



\begin{figure}[t]
\centering
\begin{algorithmic}[1]
\STATE {\bf Input}: Original lexicon as $L_0$
\STATE Train G2P model using $L_0$ and generate lexicon $L_1$ for OOV words
\STATE Merge lexicons $L_0$ and $L_1$ to train GMM-HMM models
\STATE Use phonetic language models to decode the training data
\STATE Use word and phone time boundaries to generate adapted lexicon $L_2$ from decoded data
\STATE Merge lexicons $L_0$, $L_1$, $L_2$ as lexicon $L^\prime$
\STATE Use lexicon $L^\prime$ to generate lattices for each training utterances
\STATE Estimate the pronunciation probability $p(w,b)$, which means the probability of word $w$ be pronounced as $b$, to maximize the data likelihood $p(O_u,w,b)$, which means the probability of word $w$ be pronounced as $b$ in utterance $O_u$
\STATE Compute the reduction of data likelihood before and after removing a specific pronunciation $b$, and prune the least reduction pronunciation
\STATE {\bf Output}: Learned lexicon $L$ with pronunciation probability
\end{algorithmic}

\caption{The procedure of the semi-supervised lexicon learning for the SEAME code-switching speech recognition}\label{fig:lex-learn}
\end{figure}

\subsection{Results}


Table \ref{tab:lex-result} reports various WER results on the two evaluation sets with different lexicons. The baseline results are obtained with the original lexicon that has the OOV rate of $\sim\!\!1.10\%$, while the remaining lexicons have zero OOV rates.  From Table \ref{tab:lex-result}, we notice that the G2P method is very effective and makes the WERS drop obviously.
Furthermore, the semi-supervised lexicon learning is beneficial, and 
further WER drops are gained over the G2P lexicons consistently.

\begin{table}[th]
  \caption{ WER (\%) results on both $eval_{man}$ and $eval_{sge}$ evaluation sets with different lexicons, the ``\#pron/word'' stands for the statistics of average pronunciations per word.}
  \label{tab:lex-result}
  \centering
  \begin{tabular}{ c c c c c}
    \toprule
    Lexicon   & \#pron/word  & $eval_{man}$  & $eval_{sge}$ \\
    \midrule
    Original Lex    &1.11   & $23.39$    & $33.04$  \\
    G2P 1-best  &1.12  & $22.85$    & $32.12$  \\
    G2P 5-best  &1.31       & $22.82$    & $31.99$  \\
    Learned Lex &1.39       & $22.76$    & $31.89$  \\
    \bottomrule
  \end{tabular}
\end{table}

\section{Semi-supervised acoustic modeling} \label{sec:semi-am}
\subsection{Motivation}
Transcribing code-switching data is challenging as it requires our transcribers to have strong multilingual expertise background and follow the ASR orthographic transcription protocol. Therefore, we are assuming there are some errors in human transcriptions. To verify our assumptions, we use our best ASR system to decode the training data and compare the decoded hypotheses with the manual transcriptions. We use the WMER to measure the quality of the human transcriptions.We find the WMER is around 10\% on the overall training data, and there are about 7 hours of data whose WMER is over 30\%. Table \ref{tab:wmer-train} reports the detail cumulative distribution of the WMER on the overall training data. 

In order to reduce the bad effect of those ``poorly'' transcribed data, one can simply
remove such a part of data during acoustic model training. However, if the data with higher WMER
can be exploited, better acoustic modeling might be yielded. From this standpoint, we 
propose to use semi-supervised training, regarding those higher WMER data as unsupervised data. 


\begin{table}[ht]
  \caption{ Cumulative distribution of the selected unsupervised data according to threshold WMER (\%)}
  \label{tab:wmer-train}
  \centering
  \begin{tabular}{ c c c}
    \toprule
    WMER (\%)    & Durations (hrs)  & Rate (\%) \\
    \midrule
    $>\!\!0$    & $67.18$    & $66.43$         \\
    $>\!\!20$   & $14.21$    & $14.05$       \\
    $>\!\!30$   & $6.99$     & $6.91$        \\
    $>\!\!40$   & $3.96$     & $3.92$        \\
    
    \bottomrule
  \end{tabular}
\end{table}

\subsection{LF-MMI based Semi-supervised training}
To fully exploit those acoustic data with higher WMER, we use the LF-MMI training method to conduct semi-supervised training as advocated in \cite{manoharsemi}.
Unlike conventional semi-supervised training methods, which uses frame-level, word-level, or utterance-level confidence scores \cite{vesely2013semi,thomas2013deep} to select supervisions, the proposed method in \cite{manoharsemi} uses the whole lattice as supervision.

We perform semi-supervised training as the following steps. First, we use the best LF-MMI acoustic models to decode those ``unsupervised'' data, obtaining lattices of alternative pronunciations. Then, the word lattices are converted to phone lattices as in \cite{povey2016purely}. Finally, we make senone supervisions from the phone lattices and do MMI training by weighting the lattice supervisions with a LM scale and the posterior of the best path from the decoded lattices.



\subsection{Results}
Table \ref{tab:semi-result} reports the WER results of the two semi-supervised LF-MMI training methods, one using the best path from the lattice as supervision, and the other using the pruned lattice as supervision. The baseline results are the best results from Table \ref{tab:lex-result}. 

From Table \ref{tab:semi-result}, several points are worth a mention.
First, when we remove the data with WMER $>\!\!30\%$, we can consistently obtain improved results compared with the baseline result where the whole training data is used for acoustic model training. This confirms our assumption that the human transcriptions of the SEAME data may have some errors.
Secondly, LF-MMI based semi-supervised training is working. It makes improvement over both the baseline system and the system trained with less data by means of the removal of the data containing errors. 
Lastly and more importantly, though poorly transcribed data can be detrimental to the LF-MMI training (from the ``Baseline" in Table \ref{tab:semi-result}), we still can exploit the data to contribute the final system by means of semi-supervised training (from the last two rows of Table \ref{tab:semi-result}).

\begin{table}[t]
  \caption{ The WER (\%) results of the semi-supervised training methods with different supervisions, where we use the acoustic  data whose WMER $> 30\%$ as unsupervised data.}
  \label{tab:semi-result}
  \centering
  \begin{tabular}{ c c c }
    \toprule
    System                      & $eval_{man}$  & $eval_{sge}$     \\
    \midrule
    Baseline                    & $22.76$       & $31.89$          \\
    Removal method              & $22.69$       & $31.66$          \\
    Best path supervision    & $22.63$       & $31.78$          \\
    Lattice supervision    & $22.57$       & $31.59$          \\
    \bottomrule
  \end{tabular}
\end{table}
 
\begin{figure}[t]
  \centering
  \includegraphics[width=\linewidth]{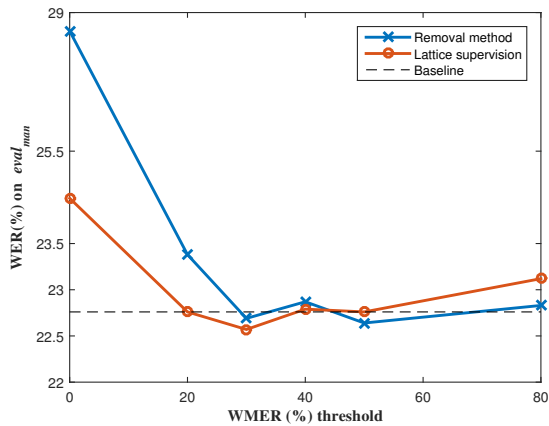}
  \caption{WER (\%) results of the lattice supervision based LF-MMI semi-supervised training on  $eval_{man}$ set using the acoustic data, whose WMER (\%) is over the specified threshold, as unsupervised data.}
  \label{fig:wer1}
\end{figure}
\begin{figure}[t]
  \centering
  \includegraphics[width=\linewidth]{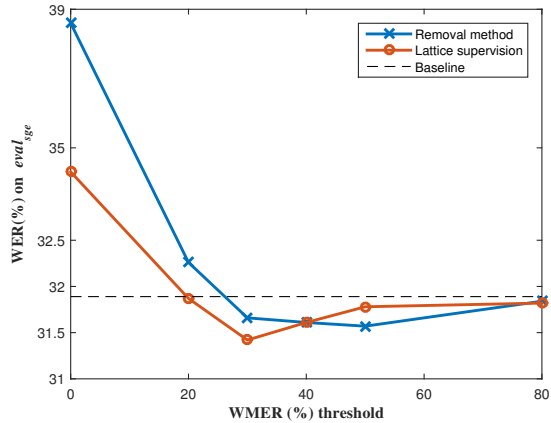}
  \caption{WER(\%) results of the lattice supervision based LF-MMI semi-supervised training on $eval_{sge}$ set using the acoustic data, whose WMER (\%) is over the specified threshold, as unsupervised data.}
  \label{fig:wer2}
\end{figure}

Moreover, to investigate the over transcription quality of the SEAME data, and the contribution of the LF-MMI semi-supervised training method, we plot Figures \ref{fig:wer1} and \ref{fig:wer2} of the WER results versus
different portion of data used as unsupervised data respectively.
From both Figures \ref{fig:wer1} and \ref{fig:wer2}, we can see the majority of the SEAME transcriptions are reliable. Either the ``removal method'' or ``lattice supervision'' based semi-supervised training yields degraded results, if we use the data whose WMER is $>\!\!20\%$ (about 14 hours of data) as poorly transcribed data. 
However, when we use the $>\!\!30\%$ data (about 7 hours of data), both two methods make improved results
compared with the baseline system. These mean the poorly transcribed data lies in $[7, 14]$ hours.  

\section{Lattice rescoring} \label{sec:lat-rescore}

Recently, it has been widely  demonstrated the effectiveness of the Recurrent Neural Network Language models (RNNLMs) \cite{mikolov2012context, xu2018neural}  over the conventional 
n-gram language models in terms of perplexity and rescoring results. 
The main advantages of the RNNLMs lie in two aspects. 
First RNNLMs can exploit much longer word history compared with the n-gram language models.
Secondly, it estimates
word probability in continuous space, alleviating data sparsity issue. 
However the drawback of the RNNLMs is that it is  hard to be employed in
online decoding due to high computational cost, and they are 
usually used in N-best or lattice rescoring.

In this paper, since we only use the training transcription to train 4-gram language models \cite{stolcke2002srilm}, it is 
a  data sparsity issue. Therefore it is worthwhile to use RNNLMs as a comparison. In practice, we use the Kaldi RNNLM toolkits \cite{xu2018neural} to train RNNLMs  and rescore the lattice. Table \ref{tab:res-result} reports our lattice rescoring results.
 From Table \ref{tab:res-result}, it can be seen that the lattice rescoring method achieves the best result with a WER of $20.54\%$ on the $eval_{man}$ data and a WER of $29.56\%$ on the $eval_{sge}$ data. 
It achieves $9\%$ and $8.8\%$ relative WER reductions respectively over the baseline system
obtained with the best semi-supervised training configuration.

\begin{table}[h]
  \caption{ WER(\%) results using lattice rescoring}
  \label{tab:res-result}
  \centering
  \begin{tabular}{ c c c }
    \toprule
    System                       & $eval_{man}$  & $eval_{sge}$  \\
    \midrule
    Best 4-gram                  & $22.57$       & $32.42$       \\
    N-best RNNLM rescoring       & $21.14$       & $30.15$       \\
    Lattice RNNLM rescoring     & $20.54$       & $29.56$       \\ 
    \bottomrule
  \end{tabular}
\end{table}

\section {Conclusions} \label{sec:con}
In this paper, we studied semi-supervised training approaches to lexicon learning and acoustic modeling under code-switching speech recognition scenarios respectively. The semi-supervised lexicon learning is to deal with the pronunciation mismatch problems. By means of collecting word pronunciation candidates and pruning them, we gained a learned lexicon that is more appropriate for the target SEAME data. As a result, we obtained better recognition results with the learned lexicon.
Furthermore, we also proposed to use LF-MMI semi-supervised training to deal with the uncertainties of human transcriptions. That is, we regard the data whose transcriptions have higher WMER as unsupervised data.
We also gained improved recognition results by the proposed semi-supervised acoustic modeling. It should be noted that there may exist similar transcribed errors in the evaluation data, we will take it into account in the future work.
Lastly to alleviate the data sparsity issue for language modeling, we employed recurrent neural network language modeling method to rescore lattice, and achieved significant word error rate reduction.



\section{Acknowledgments}
The research work is supported by the National Key Research and Development Program of China (Grant No.2017YFB1002102) and the National Natural Science Foundation of China (Grant No.61571363).

\clearpage

\bibliographystyle{IEEEtran}

\bibliography{mybib}

\end{document}